# AI Exposure Scores:

## What they measure, what they miss, and what comes next

Campbell Lund[1], Thomas Euyang[1], Zanele Munyikwa[1], and Marzieh Fadaee[1]

[1]Cohere Labs

Corresponding authors: {campbell.lund, zanele.munyikwa, marzieh}@cohere.com

## Summary

A set of exposure scores calculated in 2023 have become a central empirical input to the future of work debate. Produced by Eloundou et al. (2023) and referenced throughout this piece as the *GPTs are GPTs* scores, they define exposure as the percentage of occupational tasks a large language model (LLM) tool can assist with. While this work is a genuine methodological contribution, as these scores travel from the time and place they were calculated, the limitations the original authors named do not always travel with them. This piece traces two gaps that have widened as a result. The first is a structural gap between what static exposure scores, such as *GPTs are GPTs*, measure and what kinds of evidence policy questions need to be reliably answered. Static scoring is, by design, backwards-looking: it captures what a specific AI system could do, against a specific occupational taxonomy, at a specific moment in time. Using the widespread diffusion of the *GPTs are GPTs* scores as a case study, we observe tangible impacts in how the temporal, geographic, and ontological limitations of these scores compound when translated into policy-facing analyses. Closing this gap is the motivation behind a growing body of work, and we survey four families of recent research that respond directly to the limitations of static exposure scoring: (1) dynamic and benchmark-based measures, (2) ensemble methods, (3) task-framework extensions, and (4) worker-centered metrics. The second gap is characteristic of the future of work debate at large, and it is the gap this piece argues needs more attention: the coordination between researchers and policymakers. The methodological work responding to the limitations of static scoring is largely siloed within the research community. The policy-relevant work—analyses which ask *who is harmed, how, and when*—continues to reference the static *GPTs are GPTs* scores without engagement with the methodological updates that would let these questions be answered more reliably. We close by asking what remains beyond better measurement. Ex-post frameworks and the deliberate, political work of reimagining what futures are worthy of building towards are additional steps towards navigating uncertainty—ones that both research and policy communities would benefit from further engagement with. We argue that closing the gap between research and policy is a shared task, with distinct but parallel responsibilities. Policymakers need to widen the evidence base they rely on, engage workers as epistemic partners, and shift the goal from prediction to preparedness. Researchers need to continue building the data infrastructure, engage with interdisciplinary and participatory methods, and produce work with the needs of policymakers in mind. Better measurement matters, but it will not close the second gap alone.

# 1 Introduction

What types of work are exposed to AI? The most widely cited estimate predicts that 80% of the U.S. workforce has at least 10% of their occupational tasks exposed to large language models (LLMs), with 19% exposed at 50% or more (Eloundou et al., 2023). In this case, "exposure" is defined as the percentage of occupational tasks a LLM tool can assist with. In more recent literature, two bodies of work have emerged on this question: those building on existing methods to ask what exposure actually means for workers, and those developing new empirical tools to measure it.

The first body is of direct interest to policymakers—it asks the questions policy actually has to answer regarding which workers need support for navigating rapid technological change. For this body, the methodological foundation is largely based upon a single paper: *GPTs are GPTs: An Early Look at the Labor Market Impact Potential of Large Language Models*, in which Eloundou et al. calculate the set of exposure scores that downstream policy-relevant research continues to build upon today. However, the further we travel from the time and place these scores were calculated, the greater the gap between what these scores can tell us and how they are being used. The second body seeks to fill these empirical gaps. Researchers are building instruments which redefine how exposure is measured over time, across geographies, and beyond the discrete-task model on which the *GPTs are GPTs* methodology relies. While these tools for measuring exposure are evolving, their findings are largely siloed within the research community, and they have not received the same level of attention as the work building directly atop the 2023 scores. The result is a growing asymmetry: the policy-relevant work which asks *who is harmed, how, and when* continues to reference the 2023 scores while the methodological updates that begin to answer these questions reliably are are largely confined to the research community.

This asymmetry between researchers and policymakers is the central concern of this piece. In an effort to speak to both groups, we organize the following sections around three moves, each addressing a different but parallel need. Our goal is to highlight newer work that is extending what the *GPTs are GPTs* scores can tell us, identify what is still missing from the future of work debate at large, and put forward recommendations to synchronize the efforts of researchers and policymakers moving forward.

The first section, *Status Quo*, examines the *GPTs are GPTs* scores as they are currently used, the limitations the original authors named, and the gaps that have widened as the scores have travelled further from the time, place, and types of questions they were designed to answer. This is the context that policymakers and researchers need in order to read and extend these findings carefully. The second section, *Where We're At*, surveys four families of recent work that directly respond to the limitations of static exposure scoring: (1) dynamic and benchmark-based measures, (2) ensemble methods, (3) task-framework extensions, and (4) worker-centered metrics. These are examples of research that focus on building out the evidence base rather than extending the scores already in circulation. We then ask a third question, *Where Are We Going and What's Missing?* This section addresses what kind of work remains beyond better measurement: ex-post governance frameworks and the imaginative and political work of articulating which futures are worthy of building toward—material that every stakeholder in the future of work debate would benefit from engaging with.

# 2 Status Quo: What Does the Most Cited Exposure Score Measure, and What Does It Miss?

The *GPTs are GPTs* scores have become an empirical foundation of the future-of-work debate. Understanding exactly what they were built to measure, and what they were not, is the first move.

## 2.1 What Do They Measure?

In 2023, Eloundou et al. first released the influential paper, *GPTs are GPTs: An Early Look at the Labor Market Impact Potential of Large Language Models*. Led by a team at OpenAI, the paper produced a set of estimates assigning each occupation in the O*NET database (National Center for O*NET Development) an "exposure" score reflecting how much of its task content could be performed faster with the assistance of a LLM tool.

The authors develop a rubric to measure the overlap between LLM capabilities and Detailed Work Activities as defined by the U.S. Department of Labor's O*NET taxonomy. Exposure is determined based on whether access to a LLM tool reduces the time for workers to complete a task by at least 50%, without reducing the quality of overall output. To grade their results, they use a combination of human and LLM judges to calculate exposure at the task level before aggregating back up to the occupation level.

In sum, these scores measure the technical feasibility of an LLM circa early 2023; against a U.S. occupational taxonomy; decomposed into discrete tasks; evaluated for time-savings on tasks with verifiable outputs. The temporal, geographic, and ontological limitations that follow are not independent of each other: a score calculated against a 2023 model, using an American taxonomy, decomposed into discrete tasks, compounds rather than simply accumulates these constraints. Eloundou et al. address these boundaries directly in the original paper, however as the scores continue to travel in both time and scope, these limitations do not always travel with them. Figure 1 maps that distance.

## 2.2 What Do They Miss?

The scores were designed to answer a specific question: *can GPT-4 perform economically valuable tasks?* That question has a reasonably tractable answer. The questions policymakers are now relying on the scores to answer—*which workers need protection, which regions are at risk, how labor markets will adjust*—are far more nuanced, and the original instrument was not designed to answer them on its own.

In the following subsections, we unpack the limitations of this methodology, directing our attention to what these scores cannot tell us about the future of work. This context is necessary for both policymakers looking to interpret analysis built on top of the *GPTs are GPTs* scores, and for researchers looking to extend them.

### 2.2.1 Temporal Limitations

Because the *GPTs are GPTs* scores are benchmarked against existing model capabilities, they inherently capture a moving technological landscape at a fixed moment. To visualize this capability gap, we use the Epoch Capabilities Index (Epoch AI)—a measure of frontier model performance across reasoning, knowledge, and task benchmarks—as a proxy for progress. This index estimates

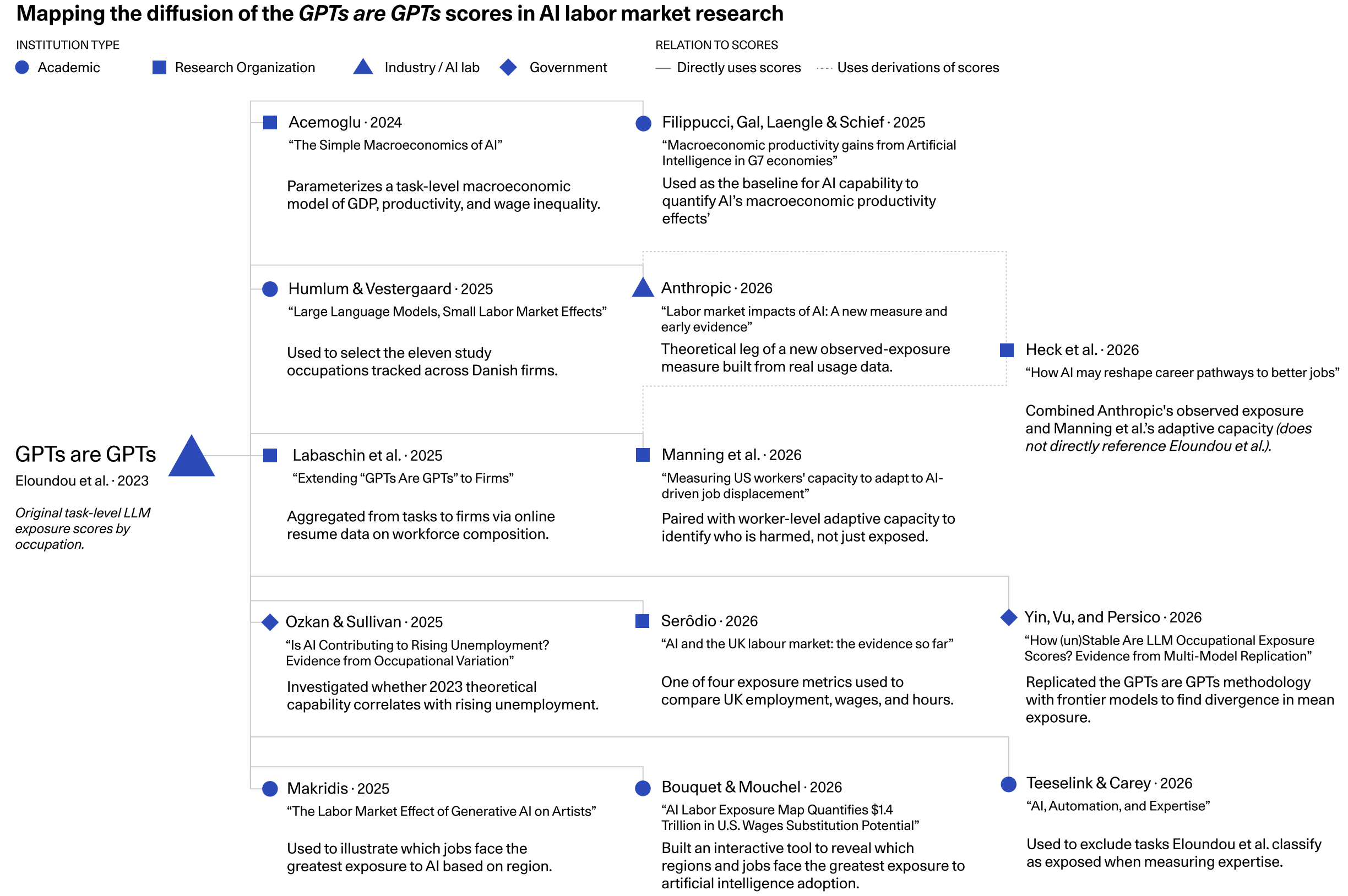

Figure 1: A non-exhaustive sample of downstream studies using Eloundou et al. (2023)'s *GPTs are GPTs* scores. Each derivative work is positioned by year (2024–2026) and shaped by institution type (academic, research organization, industry/AI lab, government). The brief annotations describe how each study uses or extends the original scores. Heck et al. (2026) is shown as a downstream node as it only references Manning et al. and Massenkoff & McCrory (2026)

a 26.0% gap between when the scores were calculated in 2023 and frontier model capabilities today (Figure 2).

This is not to say that capability correlates with economic impact—not only do the effects of a technology lag well behind the capability frontier, but the question of whether a business adopts a given technology is far more complicated than how capable it is (Brynjolfsson et al., 2021). But, unlike structural or causal models, which are designed to reason forward from underlying mechanisms, the *GPTs are GPTs* scores function as fixed snapshots: they capture what a specific model could do at a specific moment, and that snapshot becomes the input to downstream analyses that may remain in use long after the underlying capabilities have moved on. To articulate this dynamic, we repurpose the words of Tyrangiel (2026):

> "To understand how fast the present is hurtling us into the future, you need a fixed point, and the fixed points are all in the past. It's like driving while looking only at the rearview mirror—plenty dangerous if the road stays straight, catastrophic if it doesn't."

While we may be able to begin closing the capability gap by updating the *GPTs are GPTs* scores to reflect the model capabilities of today, they will always be signals, not forecasts. A new study by, Yin et al., tests this hypothesis—replicating the exact rubric from Eloundou et al. with three 2026 models (ChatGPT-5, Gemini 2.5, and Claude 4.5). They find a a 3.6-fold divergence in mean exposure, suggesting not only a capability gap, but that the rubric itself is calibrated specifically for the GPT-4 model used. Their conclusion: "Until the field develops exposure measures whose properties are stable across rating instruments, or adopts multi-annotator sensitivity as a reporting standard, the rapidly growing empirical literature on AI and the labor market rests on a measurement foundation that is itself a function of the technology it seeks to evaluate."

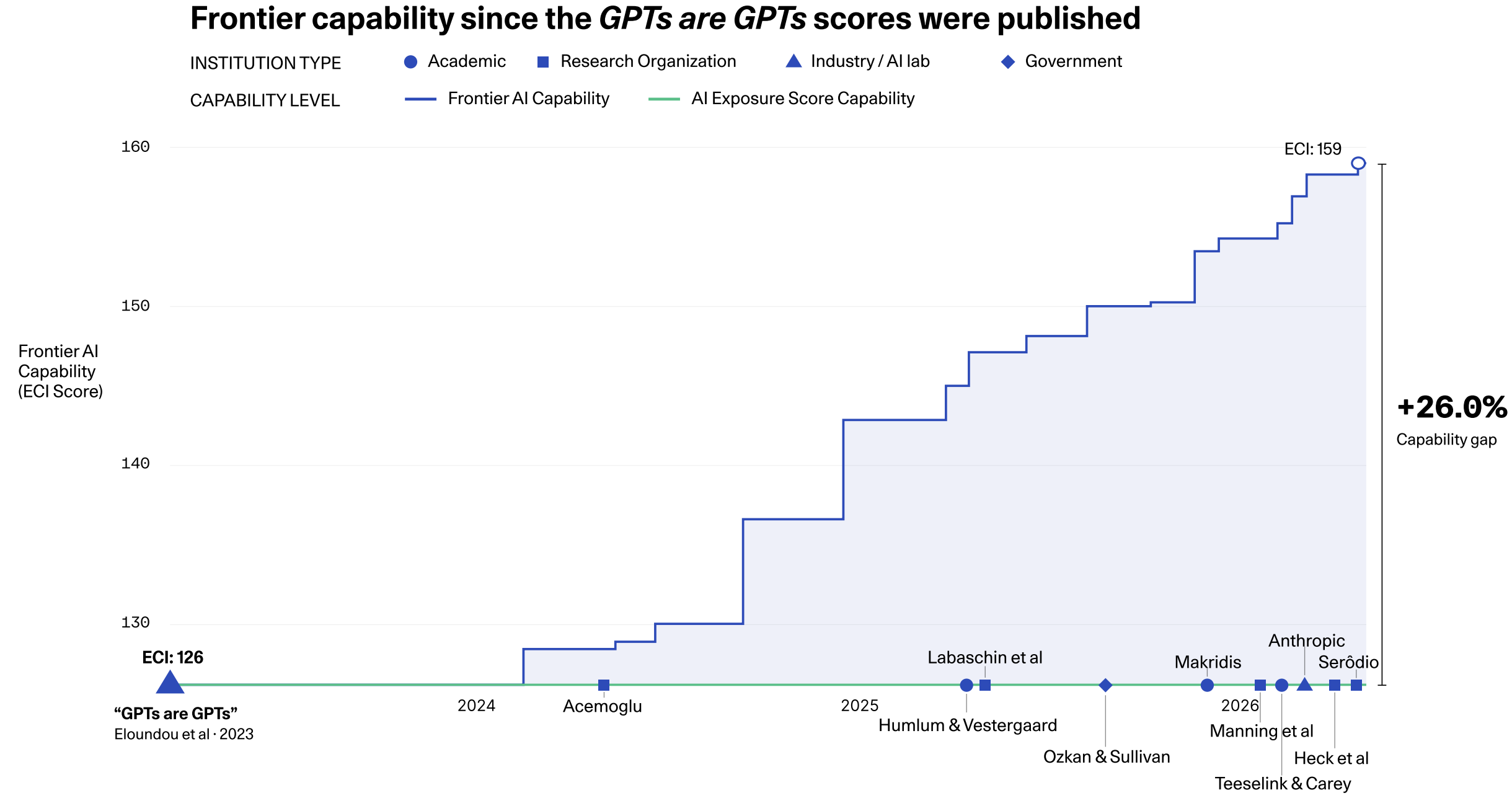

Figure 2: The widening gap between frontier model capability and the model capability the *GPTs are GPTs* scores were calibrated against. The upper line traces frontier AI capability over time, using the Epoch Capabilities Index (ECI) as a proxy (Epoch AI). The lower line marks the ECI value at the time Eloundou et al. (2023) calculated their scores, against GPT-4 (ECI: 126). Each downstream study referenced in this piece is plotted at its publication date along the lower line—showing the gap between the capability of frontier models and the capabilities assessed in the downstream study. Over the three years covered, frontier capability has risen by approximately 26.0%.

Beyond instrument sensitivity, evaluating exposure on a fixed taxonomy of work does not account for how AI may alter which tasks exist, restructure workflows, or create categories of work not present in the taxonomy at the time the scores are calculated (Merola et al., 2026; del Rio-Chanona et al., 2025).[1] The road from better signals to better forecasts is marked by measurement instruments that can be tracked over time—signals that are updated, versioned, and validated rather than fixed at a single point.

### 2.2.2 Geographic Limitations

The *GPTs are GPTs* scores are calculated against the U.S. Department of Labor's O*NET database—a taxonomy built to describe the American labor market, in American workplace terms, for American policy purposes (National Center for O*NET Development). Because of this, the scores cannot be separated from American workplace contexts, which matters for a debate that is, by necessity, global.

[1] The O*NET database itself illustrates the evolution of work to a small extent: across twelve quarterly updates since 2023, tasks have been re-rated, dropped, and added. We track these changes in a companion tool to explore how an occupation's tasks have shifted over time: `https://task-evolution.vercel.app/`

First, the scores are limited by language: O*NET task descriptions are written in English for American workplace contexts, and extending the scores internationally requires translation or assumed equivalence—both of which introduce systematic distortions. LLM graders carry documented performance gaps across languages, meaning the capability assessment itself is language-biased, not just the taxonomy it draws from.

The types of occupations represented within O*NET tell another story of U.S.-centrism. Notably, the dataset contains no discrete categories for data workers—the actual labor that powers AI systems and is often offshored from the U.S. to the Global South (Du & Okolo, 2025). The labor of these workers is structurally embedded in every LLM whose capabilities O*NET tasks are being evaluated against, yet they remain outside the policy debate those scores enable.

Finally, the taxonomy embeds assumptions about how work is formally structured. Scaling any exposure score which relies on O*NET globally does not account for the fact that occupational tasks, degrees of formalization, and education levels differ substantially by region (Merola et al., 2026). Together, these limitations shape how research outside of the U.S. is able to use the scores at all.

| **Study** | **Region** | **Use of the *GPTs are GPTs* scores** | **Crosswalk required** | **Effect on the analysis** |
|---|---|---|---|---|
| Humlum & Vestergaard (2025) | Denmark | To identify which occupations to track in a study of AI chatbot adoption across Danish firms | U.S. SOC → ISCO | Restricts the analysis to the subset of occupations where a clean correspondence between the two taxonomies exists; the American taxonomy narrows what can be examined |
| Serôdio (2026) | United Kingdom | As one of four exposure measures in an analysis of UK employment, wages, and hours | U.S. SOC → UK SOC2020 | Adds a translation layer, with its own assumed equivalences, between the scores and the labor market being described |

Table 1: Two recent non-U.S. studies illustrate how the *GPTs are GPTs* scores travel across geographic contexts. Each had to bridge from the U.S. occupational classification O*NET is built on (SOC) to a different national or international taxonomy—and each pays a cost for doing so. Humlum & Vestergaard recognize that their analysis is narrowed by the requirement; Serôdio recognizes the limitations with assumed equivalence.

### 2.2.3 Ontological Boundaries

The validity of the *GPTs are GPTs* scores rests on the premise that work can be decomposed into a discrete bundle of tasks. As the authors themselves acknowledge:

> "It is unclear to what extent occupations can be entirely broken down into tasks, and whether this approach systematically omits certain categories of skills or tasks that are tacitly required for competent performance of a job [...] If indeed, the task-based breakdown is not a valid representation of how most work in an occupation is performed, our exposure analysis would largely be invalidated." (Eloundou et al., 2023)

Tacit knowledge which cannot be fully articulated, relational work, and the importance of workflow are all essential elements of work which are not factored into this calculation. An isolated list of tasks captures what can be itemized; it cannot capture the connective tissue between items. Further, relying on human or LLM judges to score task success limits the types of tasks which can be assessed. What Eloundou et al.'s methodology does not account for are the elements of work that involve judgment, trust, and relationships which are not scorable outside of specific workplace contexts.

Evidence from Yin et al. (2026) begins to support this claim. After replicating Eloundou et al.'s methodology across three frontier models, they find that cross-model disagreement is not uniformly distributed across the occupational structure. The largest disagreements concentrate in supervisory roles and occupations that combine cognitive and physical tasks—the same categories of work in which tacit knowledge, relational work, and judgment are consequential.

Where the task-based lens is clearest (purely physical work, purely linguistic work) the rubric produces stable results. Where it is most strained (work that combines judgment with procedure, or relational context with codifiable output) the rubric produces its largest instabilities. The occupations where policymakers most need reliable estimates are, by this logic, the occupations where the task-based decomposition is least equipped to provide them. Thus the scores do not reveal a natural fact about which work is automatable. They reveal which work *appears* to be automatable when work is viewed through a particular, task-based lens—a different lens produces a different distribution of exposure.

## 3 Where We're At: Bridging the Empirical Gaps

If the *GPTs are GPTs* scores are able to support specific questions about task-level exposure, what would a more complete ex-ante picture look like? Here we shift our focus to the second body of work: those developing new empirical tools for measuring exposure.

In this section, we provide examples of four different approaches: (1) benchmark-based and dynamic measures, (2) ensemble methods, (3) extensions to the task-based framework, and (4) worker-centered measures. These examples are of direct interest to the research community as some of the most innovative methods for understaniding the impacts of AI on the labor market, but are also useful for any policymaker looking to understand the technical work behind improving labor market signaling. None of these approaches discard exposure scoring—some use the *GPTs are GPTs* scores as their starting point. Together, they extend what those scores can tell us.

### 3.1 Benchmark-Based and Dynamic Measures

To directly address the temporal limitations, ex-ante measurements should be able to reflect updating advances in AI capability. Dominski & Lee (2025) construct one example of this: a dynamic occupational exposure index evaluating current frontier models at five stages of capability—from narrow ML through agentic AI. Going one step further, they link the resulting scores to Current Population Survey data for a near real-time labor market analysis. Comparing late 2022 through early 2023 against late 2024 through early 2025, they find effects across multiple dimensions of labor market impact: highly-exposed occupations have seen reductions in employment, increased unemployment, and shorter work hours, with effects concentrated in complex reasoning and problem-solving tasks. In their regression analysis, a 10-point increase in occupation-level exposure is associated with a 5.6 to 8.5 percentage point decline in employment, 0.31 fewer hours at the main job, and a 0.93 percentage point decline in the full-time employment share. This is among the first empirical evidence that exposure, as measured dynamically, predicts observable labor market outcomes and not just theoretical susceptibility.

The Remote Labor Index (Mazeika et al., 2025) takes a complementary approach. Rather than measuring exposure through expert or model annotation of task descriptions, the index benchmarks actual AI agent performance on real-world remote work tasks drawn from genuine professional contexts. At publication, frontier agents completed only 2.5% of sampled tasks acceptably, providing a concrete baseline that updates as model capabilities change. The contribution here is distinct from Dominski & Lee's: where they link dynamic scores to observed labor market outcomes, the Remote Labor Index tracks what AI systems can demonstrably do today—not what annotators estimate they could do. This approach advances the measurement of exposure itself rather than its downstream effects.

## 3.2 Ensemble Approaches

The fragility of a single exposure score can be strengthened by an ensemble approach. Frank et al. (2025) demonstrate this: drawing on a decade of exposure measures across studies, they find that individual scores are weakly correlated and sometimes anti-correlated with one another, and that no single score predicts unemployment risk well on its own. They show that an ensemble combining five established exposure frameworks (Frey & Osborne, 2017; Brynjolfsson et al., 2018; Felten et al., 2018; Eloundou et al., 2023), weighted to emphasize the most informative scores, performs substantially better. The ensemble explains 29.8% of variation in occupation-level unemployment risk on its own and contributes an additional 18 percentage points beyond what education, skills, and regional controls already explain, for 75.5% of total variation in the full model. Because individual scores are weakly or even negatively correlated with each other, the combination captures dimensions of exposure that no single score can reach. Their conclusion is direct: "...efforts using only one AI exposure score will misrepresent AI's impact on the future of work." For policymakers, this finding has a practical implication: relying on a single score, however widely cited, is an analytically fragile choice.

## 3.3 Extensions to the Task-Based Framework

The ontological limitation of treating work as a discrete bundle of tasks is addressed by two recent papers.

Demirer et al. (2026) focus on the relation between tasks—they model production as a sequence of interdependent steps rather than separable units. AI executes contiguous chains of steps, and firms then re-bundle the remaining steps into new jobs, trading off specialization against coordination. A score that treats tasks as independent will systematically misread which jobs are at risk: it is the adjacency and sequencing of tasks, not just their ability to be performed by an AI tool in isolation, that determines susceptibility to automation. They show that exposure scores can be supplemented with information about task interdependencies—data which job postings and workflow analyses can begin to provide.

Klein Teeselink & Carey (2026) focus on the equivalence of tasks. Using hundreds of millions of job postings across 39 countries, they distinguish expertise-raising from expertise-lowering automation and find that only the latter predicts wage stagnation and job loss. High-exposure occupations saw a 6.1% decline in postings overall, with effects suggestively larger in countries with stricter employment protection legislation and suggestively smaller in countries with greater digital readiness (though the authors note limited statistical power with only 39 countries in the sample). Their conclusion: "...exposure-based measures alone obscure economically meaningful heterogeneity." Notably, their analysis uses Eloundou et al.'s $\beta$ scores as its primary measure of occupation-level exposure, illustrating that exposure scores are a meaningful foundation for differentiated analysis, even as their findings complicate what aggregate exposure can tell us. Which tasks get automated matters as much as how many, and standard task-based exposure scores cannot distinguish between the two.

### 3.4 Worker-Centered Measures

The approaches above begin to address what AI can do, how exposure travels through occupational networks, and how task structure shapes outcomes. What they do not address is the worker—neither what workers want, nor how prepared workers are to navigate technological change. This is when asking the question, *by whom and according to whom*, becomes consequential: measurement frameworks that exclude worker preferences will paint a picture of what technology can do, not what technology should do. Exposure scores capture technical feasibility, but worker-centered measures represent which tasks workers do not want automated, and which workers are least equipped to absorb displacement.

Shao et al. (2026) begin to address the question of what AI should do. They audit 844 tasks across 104 occupations with input from 1,500 domain workers, pairing AI capability assessments with worker desire and a Human Agency Scale that captures workers' preferred levels of human involvement in each task. The result is four zones: tasks workers and researchers agree should be automated; tasks workers do not want automated regardless of technical feasibility (the "Red Light" zone); tasks where capability lags desire; and tasks where neither supports automation. The Red Light category matters most for policy because it identifies cases where technical exposure overstates likely adoption—not because the technology is incapable, but because workers, for a range of reasons, do not want it automated. This example is just one step towards including workers in the debate. For researchers, further engagement with workers as epistemic partners is necessary in order to expand the perspectives and types of preference represented (Miceli et al., 2025). Ultimately, a score that registers exposure without registering desire will systematically over-predict displacement in domains where workers retain agency over how their work is done.

Manning et al. (2026) bring workers' adaptive capacity into the picture. They pair exposure estimates with an adaptability index measuring workers' capacity to respond to AI-driven displacement. Considering factors such as financial buffers, transferable skills, geographic mobility, and age, they find broad resilience across the U.S. labor market overall, but concentrated pockets of vulnerability in specific occupational groups and regions where exposure is high and adaptive capacity is low. For policymakers, this view is more actionable than exposure alone: it identifies not just where AI pressure is highest but where the workforce is least equipped to respond, which is essential for designing transition support, reskilling programs, and targeted protections. Critically, Manning et al. also find that adaptive capacity and exposure are positively correlated: highly exposed occupations tend to be held by financially secure, skilled, and well-networked workers who are relatively well-positioned to adapt. Exposure alone does not map cleanly onto vulnerability.

Taken together, these four approaches share a common logic: exposure scores as a starting point, not a final answer. The methodological direction is not toward refining a single number but toward building a more useful measurement infrastructure—one that, if done carefully, improves fidelity and confidence and produces evidence that is more actionable for the decisions policymakers actually face.

# 4 Where We're Going and What's Missing: How to Navigate the Unknown

The four approaches above collectively begin to address the temporal, geographic, and ontological limits of a single exposure score—more research is necessary across each of these dimensions in order to continue sharpening our empirical tools. But these are not the only tools in our toolkit for preparing for the future of work. Even when utilized together, they share a common constraint: they are tools for anticipating risk, not for responding to real-world impacts as they actually materialize. They also do not inherently help us imagine what the futures we are building towards should look like. Two further moves are necessary for both researchers and policymakers: a shift from prediction to preparedness, and a more deliberate engagement with the political work of imagining and advocating for the futures we want.

## 4.1 Moving Beyond Ex-Ante Evidence

Exposure scores can tell us where to look—which occupations and groups are most likely to feel the effects of advancing AI—but they cannot tell us with certainty what those effects will be, when they will arrive, or how they will propagate through complex occupational networks and labor markets. So how do we prepare for the future?

In the context of AI governance, ex-post frames are about responding to risks after the fact, rather than attempting to fully anticipate them in advance. Marchant & Stevens (2017) characterize the broader risk-governance landscape as offering four approaches to emerging technology: two ex-ante (risk analysis and precaution) and two ex-post (liability and resilience). They argue that traditional ex-ante approaches—specifically risk analysis—struggle with anticipatory governance because of the difficulty of ascertaining downstream consequences. For the same reason AI exposure scores are limited forecasters, governance which relies only on quantified risk may not account for the ways in which emerging technologies will continue to alter the world we live in. Resilience, by contrast, asks not what risks will materialize but what institutions, protections, and capacities help workers and communities navigate technological change regardless of which specific predictions hold.

Applied to AI's impact on the labor market, the resilience frame shifts the emphasis from quantifying risk to absorbing shock. Narayanan & Kapoor (2025) argue that the most defensible policy moves are those whose value does not depend on getting any particular forecast right. They emphasize "no regret" policies such as: protecting the foundations of democracy and the free press, supporting equitable labor markets, building technical and institutional capacity in government, investing in early-warning systems and adverse-event reporting, and strengthening worker protections. These are interventions that manage risk regardless of how AI capabilities and adoption unfold over the coming decade, because they invest in the institutional infrastructure that any version of the future will require.

For policymakers, this reframe means that policy should be designed to be flexible and not dependent on a static measurement or single set of exposure scores. Scenario-based policy with clearly identified triggers for activating different responses is more robust to forecasting error than policy anchored to a forecast.

## 4.2 Working for the Futures We Want

Ex-ante metrics can make us feel like the future is out of our control. Ex-post frameworks can feel like preparing for the worst. Neither inherently empower us to imagine and advocate for the futures we want.

The question of what we want the future to look like requires a different kind of work than measurement. Benjamin (2024) frames this as the work of imagination—not as a private creative exercise but as a collective political one. Imagination is the deliberate effort to envision the worlds we want to build, and a precondition for refusing the worlds we are being told are inevitable. This framing insists that other trajectories are possible and that the question of which trajectory we pursue is a political one.

A score telling us that 80% of the U.S. workforce has at least 10% of their tasks exposed to large language models describes technical feasibility under a particular set of assumptions—not a forecast, and not a mandate. Acemoglu et al. (2026) further emphasize this point, arguing that the realized impact of AI on workers is a function of incentives, market failures, and policy choices—not of technological determinism. Along our current trajectory, they identify a pervasive pro-automation bias in how AI is being developed and deployed, driven by misaligned firm and developer incentives, not by any inherent property of the technology. Their framework distinguishes five categories of technological change, only one of which—new task-creating AI—is unambiguously pro-worker.

However, the dominance of automating applications over augmenting ones is a direction that policy can actively reshape, and workers can reimagine (Duarte et al., 2025). How exposure translates into outcomes for workers depends on decisions about how AI is built, deployed, and governed that are still being made. Some workers have already begun imagining and articulating their visions for the future, along with recommendations for how to start building it (AFL-CIO, 2025). But Reid Hoffman, co-founder of LinkedIn and a Microsoft board member, has observed that external pressure for AI returns leads many CEOs to treat layoffs as inevitable: "A lot of them have convinced themselves this only ends one way. Which I think is a failure of the imagination" (Tyrangiel, 2026).

It is worth clarifying that the work of imagination is not the work of converging on a single shared future. What a desirable future looks like is context-specific and shaped by many (sometimes competing) interests and constituencies. The first step toward reconciling competing visions is being able to articulate what each constituency wants and why, and to make those articulations available to deliberation rather than letting them be settled by larger structures of power.

The path forward, then, is not exposure scores plus better empirical evidence alone. It is exposure scores plus ex-post resilience frameworks plus the deliberate, collective work of articulating and advocating for the futures we want (Figure 3). As Benjamin (2024) puts it, imagination is not a luxury—it is a necessity. Knowing what AI could do to work is only useful if we have a clear view of what we want work to become.

# 5 Recommendations

Building the evidence base we call for throughout this piece is not only an academic challenge—it is an institutional one. AI systems are increasingly being deployed across workplaces, public services, and enterprise settings worldwide, yet the empirical infrastructure for understanding their labor-market effects remains fragmented, U.S.-centric, and slow-moving. More open, reproducible, and globally representative research is necessary if policymakers, researchers, and organizations are to make coordinated and informed decisions under conditions of rapid technological change.

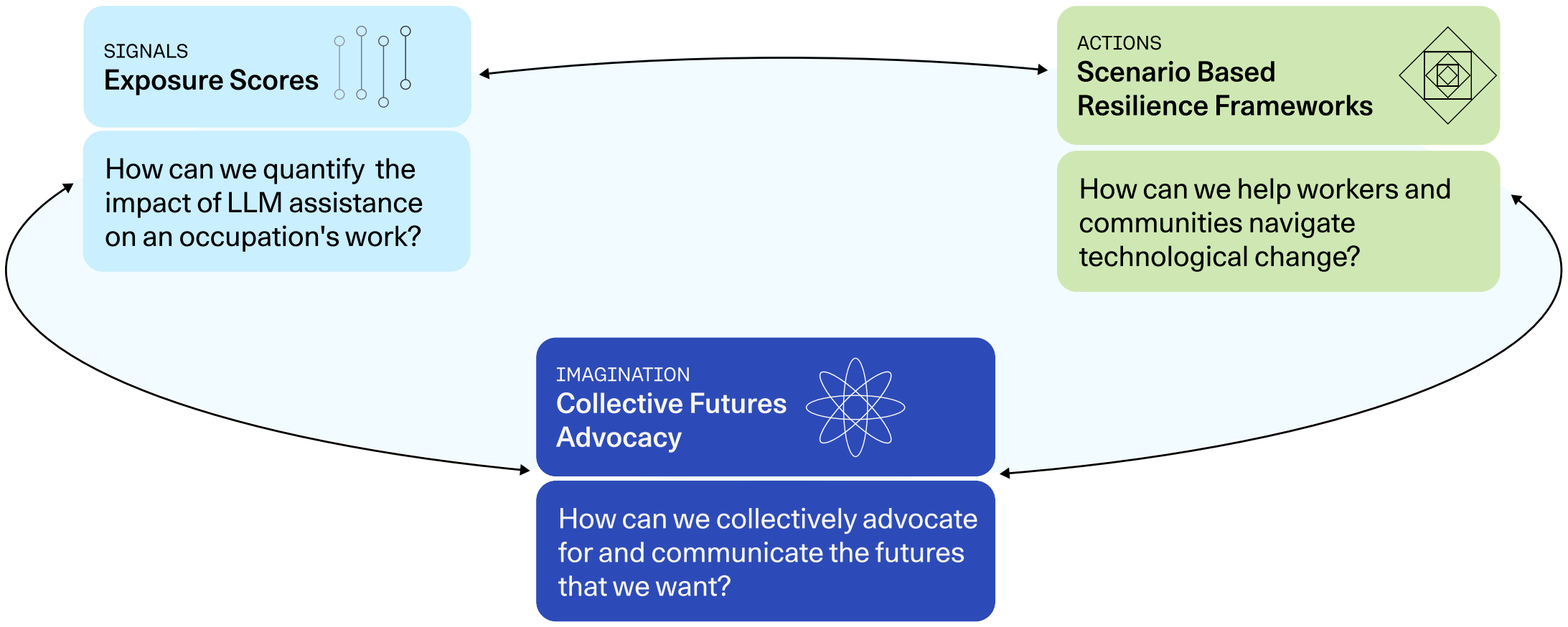

Figure 3: Three complementary modes of engaging with the future of work: signals (exposure scores and other ex-ante measurements), actions (scenario-based resilience frameworks for navigating uncertainty), and imagination (collective advocacy for the futures we want to build). Each mode answers a different question, and none is sufficient on its own.

Two groups in particular can take actionable steps to this end—policymakers and researchers. The responsibility of policymakers is to develop plans to protect workers in the face of uncertainty. Exposure scores can be interpreted as a signal for identifying who might be vulnerable, but they should not be relied on as a map of the future. The responsibility of researchers is parallel: to continue developing and updating empirical instruments with the same rigor applied to any other measurement tool—explicit about scope, versioned as capabilities change, and validated against the questions they are actually designed to answer.

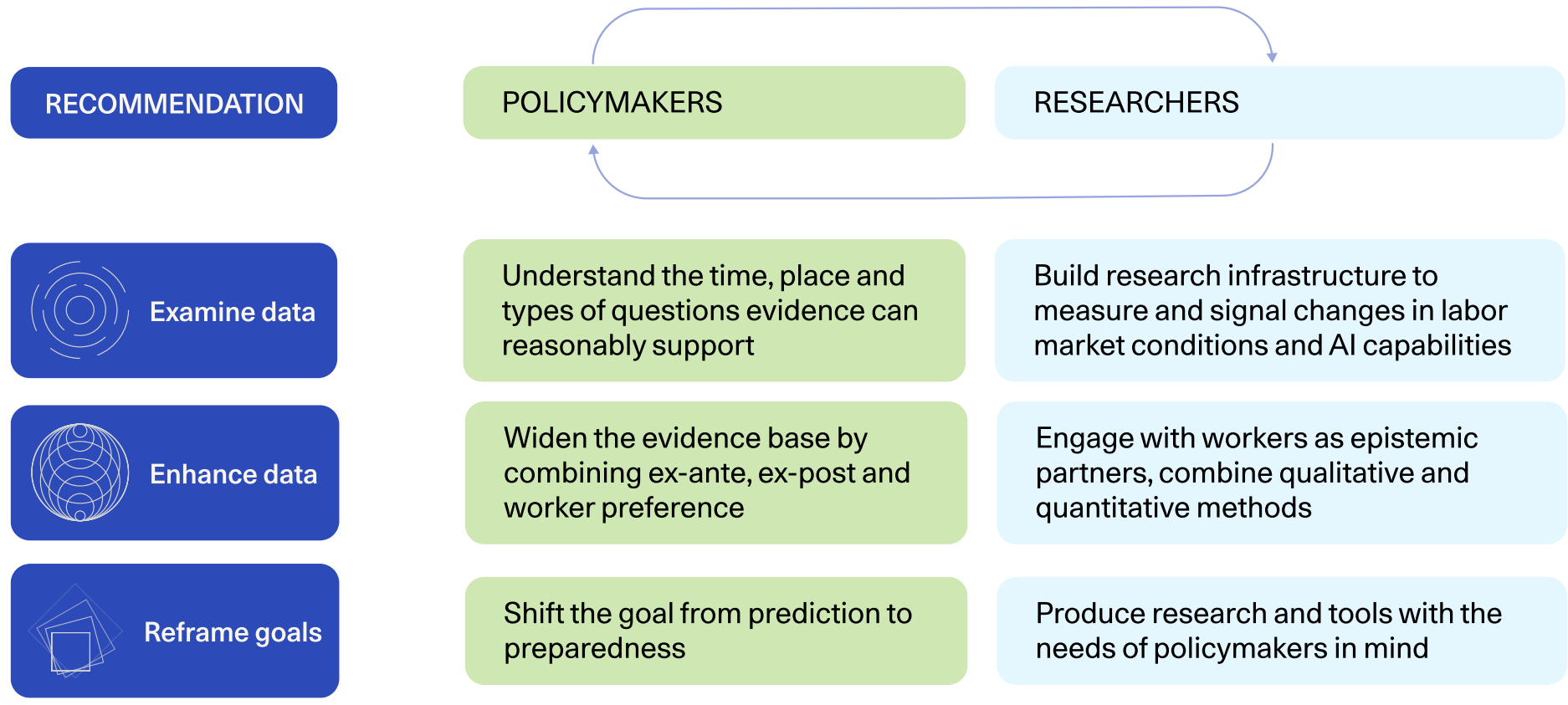

Figure 4: A summary of recommendations for policymakers and researchers, organized around three shared moves: examining the data, enhancing the data, and reframing the underlying goals. Each row pairs a policymaker recommendation with the corresponding researcher recommendation, illustrating how the two sets of practices reinforce one another. The full text of each recommendation appears in the Recommendations section.

## 5.1 Recommendations for Policymakers

1. **Understand the scope of ex-ante evidence.** A score or exposure prediction often measures technical feasibility, at a moment in time, against a particular taxonomy. As such, these scores need to be continuously updated in order to paint a full picture of the current AI capability landscape, and should not be extended beyond their intended scope. *To read more, see: Merola et al. (2026); del Rio-Chanona et al. (2025) Yin et al. (2026).*

2. **Expand the evidence base.** Pairing exposure scores with other forms of evidence—dynamic measures, ensemble methods, task-framework extensions, and worker-centered metrics—helps to expand the types of questions ex-ante evidence can support. Jurisdictions across the world should not rely on scores built from other countries' taxonomies without explicit validation for the relevant context. Workers who are not included in these taxonomies deserve particular analytical and policy attention. *To read more, see: Dominski & Lee (2025); Mazeika et al. (2025); Frank et al. (2025); Demirer et al. (2026); Klein Teeselink & Carey (2026); Shao et al. (2026); Manning et al. (2026).*

3. **Engage with workers as epistemic partners rather than as subjects of analysis.** Workers have direct knowledge of how their tasks are changing, which elements of their work are not captured in any taxonomy, and what kinds of change they find acceptable. Policy built without that input will systematically exclude situational knowledge and worker preference *To read more, see: AFL-CIO (2025); Mateescu et al. (2026).*

4. **Shift the goal from prediction to preparedness.** Frameworks built around predicting which jobs will be exposed make policy success contingent on the predictive accuracy of any single set of scores. A resilience-oriented posture asks instead what institutions, protections, and capacities help workers and communities navigate technological change — regardless of which specific predictions hold. Scenario-based policy with clearly identified triggers for activating different responses is more robust to forecasting error than policy anchored to a forecast *To read more, see: Narayanan & Kapoor (2025); Marchant & Stevens (2017).*

## 5.2 Recommendations for Researchers

1. **Continue building the data infrastructure.** There is a lack of adequate and timely data for measuring AI's impact on the labor market, and the most useful research interventions are those that work to close this gap rather than refine existing projections. Improvement can be made regarding data coverage (who is included), granularity (shifting from isolated task-level exposure to analyzing workflows, firms, sectors, and networks of skill overlap), and frequency (measures which update alongside AI capabilities). *To read more, see: National Bureau of Economic Research.*

2. **Engage with interdisciplinary and participatory methods.** Single-discipline approaches will not produce rich enough evidence to answer socio-technical questions. The questions at stake — how AI will affect work, for whom, under what conditions, and with what implications for equality and agency — are not purely technical. They require qualitative and ethnographic methods alongside quantitative ones, and participatory approaches that engage workers and affected communities as researchers with epistemic authority over their own labor conditions, not merely as data sources *To read more, see: Miceli et al. (2025); Lange et al. (2026).*

3. **Produce research to directly inform policy.** Policy decisions require knowing not just whether AI will affect a given occupation, but when a given response should be activated and what kind of evidence would justify that activation. This means moving from static exposure estimates toward indicators that policymakers can monitor over time: early warning signals, threshold triggers, and ex-post assessments of whether predicted effects have materialized. The goal is research that is decision-relevant, not just empirically interesting. *To read more, see: Comunale & Manera (2024).*

## 6 Conclusion

This piece has examined what exposure scores can and cannot tell us, and surveyed the emerging body of work building more precise empirical tools to fill those gaps. We use the *GPTs are GPTs* scores as a case study to illustrate larger observations about the future of work debate: the disconnect between research and policy agendas, and the need for more types of evidence across all fronts. Karger et al. (2026) note that current debates over AI's economic impact often try to answer three distinct questions in a single breath:

> "First, will AI capabilities advance meaningfully, such that AI systems become capable of independently performing, or assisting with, a large quantity of economically valuable work?
>
> Second, if such progress occurs, what will happen to key macroeconomic outcomes, including GDP growth, productivity, labor-force participation, and inequality?
>
> And third, given predictions and uncertainty about the effects of AI on the economy, what are the optimal policy responses?"

These are related but not equivalent questions, and answering the first does not automatically illuminate the second or third. Task-based exposure scores were designed to answer a version of the first. The derivative work we surveyed extends them toward the second. But the third—the question policymakers actually have to answer—cannot be inferred from the first two alone. It requires combining ex-ante measurement with ex-post preparation, with research on adoption and adjustment, with deliberation on which futures of work are worth building towards.

The concern this piece has traced is not with exposure scoring as a practice, but with the distance between the needs of policymakers and current research agendas. Minimizing that distance is about improving the coordination between the two: building evidence that is rigorous, current, globally representative, and building the channels through which that evidence actually informs the decisions being made. At Cohere Labs, we are committed to following the recommendations we put forth for researchers as we continue to explore the unknown future of work, together.

## Acknowledgments

We want to thank the many brilliant minds who came together from across disciplines to share their expertise throughout the process of producing this report. In particular, thanks to, *Joelle Pineau, Halak Shrivastava, A.J. Bhadelia, Julia Kligman, and Ella Morley* for all your reviewing.